\documentclass[letterpaper]{article}

\usepackage{aaai2026}
\nocopyright 

\usepackage{times}
\usepackage{helvet}
\usepackage{courier}
\usepackage[hyphens]{url}
\usepackage{graphicx}
\usepackage{natbib}
\usepackage{caption}
\usepackage{enumitem}
\usepackage{algorithm}
\usepackage{algorithmic}

\usepackage[utf8]{inputenc}
\usepackage{amsmath,amssymb,amsfonts}

\usepackage{booktabs}
\usepackage{subcaption}
\usepackage{placeins}
\usepackage{microtype}
\usepackage{tabularx}

\newcommand{\vecm}{\mathbf{m}}

\title{Plato's Cave: A Human-Centered Research Verification System}

\author{
Matheus Kunzler Maldaner\textsuperscript{\rm 1}\thanks{These authors contributed equally, ordered by date of birth.},
Raul Valle\textsuperscript{\rm 1}\footnotemark[1],
Junsung Kim\textsuperscript{\rm 1},
Tonuka Sultan\textsuperscript{\rm 1},\\
Pranav Bhargava\textsuperscript{\rm 1},
Matthew Maloni\textsuperscript{\rm 1},
John Courtney\textsuperscript{\rm 1},
Hoang Nguyen\textsuperscript{\rm 1}, \\
Aamogh Sawant\textsuperscript{\rm 2},
Kristian O'Connor\textsuperscript{\rm 1},
Stephen Wormald\textsuperscript{\rm 1},
Damon L. Woodard\textsuperscript{\rm 1}
}

\affiliations{
\textsuperscript{\rm 1}University of Florida\\
\textsuperscript{\rm 2}Georgia Institute of Technology\\
\texttt{\{mkunzlermaldaner, rvalle1\}@ufl.edu}}

\newcommand{\numPapersProcessed}{104}

\newcommand{\numDAGSamplesPerPaper}{8}   
\newcommand{\numNodeWeightSamples}{8}    
\newcommand{\maxNodesPerDAG}{16}

\newcommand{\numDAGSamplesTotal}{832}    
\newcommand{\numDAGSamplesValid}{810}
\newcommand{\numDAGSamplesInvalid}{22}

\newcommand{\numTrialsTotal}{6480}       

\newcommand{\modelDAG}{\texttt{gpt-5-mini}}
\newcommand{\modelNode}{\texttt{gpt-5-nano}}

\newcommand{\wallclockHours}{13.99}
\newcommand{\paperRuntimeMedianMin}{65.4}

\newcommand{\costAPI}{0.15}

\newcommand{\mkm}[1]{
}

\newcommand{\hide}[1]{}

\usepackage[most]{tcolorbox}
\newtcolorbox{promptbox}[1]{
  enhanced,
  breakable,
  width=\linewidth,
  colback=white,
  colframe=black!100,
  boxrule=0.8pt,
  arc=10pt,
  left=8pt,right=8pt,top=4pt,bottom=4pt,
  before skip=8pt,
  after skip=8pt,
  fontupper=\ttfamily\footnotesize,
  attach boxed title to top left={
    xshift=12pt,
    yshift*=-\tcboxedtitleheight/2
  },
  fonttitle=\bfseries\footnotesize,
  boxed title style={
    colback=black!60,
    colframe=black!100,
    boxrule=1pt,
    arc=5pt,
    left=5pt,right=5pt,top=2pt,bottom=2pt
  },
  title=\strut#1
}

\date{}

\begin{document}

\maketitle

\begin{abstract}
The growing publication rate of research papers has created an urgent need for better ways to fact-check information, assess writing quality, and identify unverifiable claims. We present Plato's Cave as an open-source, human-centered research verification system that (i) creates a directed acyclic graph (DAG) from a document (ii) leverages web agents to assign credibility scores to nodes and edges from the DAG and (iii) gives a final score by interpreting and evaluating the paper's argumentative structure. We report the system implementation and results on a collected dataset of 104 research papers.
\end{abstract}

\section{Introduction}\label{sec:intro}

Scientific publishing continues to scale faster than traditional review capacity. In 2022, there were 3.3 million research publications in science and engineering \cite{NSB2023} and in the field of artificial intelligence, this pressure has been increasing. Popular machine learning conferences such as NeurIPS, ICML and ICLR received more than 60,000 combined submissions in the last cycle alone \cite{neurips2025blog, papercopilot2026}. This proliferation has strained traditional peer review mechanisms and allowed the spread of unreliable research. The total number of articles in the Retraction Watch Database increased 26\% in 2025, now exceeding 63,000, and high-profile cases of fabricated data have undermined confidence in published findings \cite{RetractionWatch2024, RetractionWatch2025}. This situation calls for the development of tools that can help researchers and reviewers rapidly audit the internal logic of papers as structured arguments where claims, methods, evidence, and conclusions are complex and interconnected.

Existing approaches towards research verification fall into three categories. (1) Citation-based metrics such as the h-index and impact factors of the journals measure popularity rather than correctness and may distort performance targets that should prioritize logic \cite{strathern:97}, (2) fact-checking systems such as SciFact \cite{wadden2020factfictionverifyingscientific} and FEVER \cite{thorne2018feverlargescaledatasetfact} operate at the sentence level and fail to capture the argumentative structure that connects claims to evidence and (3) approaches to pure language models lack an in-depth knowledge of external sources and cannot reliably verify claims against primary sources \cite{liu2023reviewergpt}.

We address these limitations by posing a system in which scientific papers are modeled as knowledge graphs, nodes represent semantic units (hypotheses, claims, evidence, methods, results), and edges encode their relationships. Our system then traverses the graph and verifies individual claims using autonomous browser-use agents that search the web to assess the credibility of sources by extracting supporting or contradicting evidence. The verified node qualities propagate through the graph structure via a trust-gating mechanism: when a parent node has low trust, it reduces confidence in all dependent child nodes, ensuring that weak foundational claims appropriately diminish confidence in their downstream linked nodes. This propagation step is inspired by message-passing inference in probabilistic graphical models (i.e. belief propagation / sum-product), but here it propagates verification-derived trust rather than probabilistic marginals \cite{pearl1988probabilistic, kschischang2001factor, yedidia2003understanding, wainwright2008variational}. 

Our contributions include:

\begin{itemize}[leftmargin=2em]
    \item A role-aware DAG representation for papers, together with a validation-and-repair step that enforces ontology and acyclicity constraints.
    \item A multi-level scoring framework that combines the quality of individual nodes, edge-level priors and semantic alignment, and global structure metrics (bridge coverage, best-path reliability, redundancy, fragility, coherence, and coverage).
    \item An efficient factorized sampling scheme that separately perturbs structure ($K$ sampled DAGs) and aggregation parameters ($M$ resampled node/propagation weights), enabling uncertainty estimates without retraining.
    \item A pilot system evaluation using $\numPapersProcessed$ research papers showing end-to-end throughput, failure modes, and a moderate alignment with coarse predetermined labels.
    \item An open-source interactive system that exposes our pipeline to the user, enabling human-in-the-loop auditing rather than replacing reviewer judgment.
\end{itemize}

The remainder of this paper proceeds as follows: Section~\ref{sec:relatedwork} reviews related work in claim verification, knowledge graphs, trust propagation, and argumentative structure in scientific documents. Section~\ref{sec:problemformulation} formalizes our approach towards the paper verification problem. Section~\ref{sec:implementation} describes our implementation including the knowledge graph ontology, multi-agent verification pipeline, and tech stack used on our system design. Section~\ref{sec:evaluation} presents the pilot evaluation and the results of the cache-first calibration. Section~\ref{sec:discussion_futurework} discusses strengths, limitations, and future work. Finally, Section~\ref{sec:conclusion} concludes.

\section{Related Work}\label{sec:relatedwork}

Research verification intersects several strands of prior work, spanning the verification of scientific claims, construction of knowledge graphs, trust propagation, and the emergence of multi‑agent systems for automated information gathering. We structure this section around these themes to situate Plato's Cave within the broader landscape of present literature.

\subsection{Scientific Claim Verification}

Automated fact-checking has received significant attention in natural language processing. The FEVER dataset introduced the task of verifying claims against Wikipedia \cite{thorne2018feverlargescaledatasetfact}, a study that led to numerous neural architectures for evidence retrieval and entailment classification. 
SciFact \cite{wadden2020factfictionverifyingscientific} adapted this paradigm to scientific claims, pairing assertions with relevant abstracts and stance labels (i.e. supports, refutes, not\_enough\_info). However, these systems operate at the sentence level and treat verification as an isolated classification task rather than modeling the argumentative structure of complete papers. 

Other studies have explored claim decomposition \cite{Liu_Datta_Lim_2015} and multi-hop reasoning \cite{yang2018hotpotqa}, but these approaches still focus on question-answering rather than holistic paper assessment. Our work differs by extracting the complete dependency graph of a specific paper's arguments and propagating trust through this structure on top of verifying claims independently.

\subsection{Knowledge Graph Construction}

Automated knowledge graph construction from text has progressed through rule-based and supervised relation \cite{CHENG202495, Liu2020}, and more recently, LLM-based approaches \cite{zhu2024llmsknowledgegraphconstruction}. Scientific knowledge graphs like Open Academic Graph \cite{10.1145/3292500.3330785}, which seeks to unify the Microsoft Academic Graph \cite{sinha2015an} and AMiner \cite{10.1145/1401890.1402008}, as well as Semantic Scholar's API \cite{Kinney2023TheSS}, focus on citation networks and metadata (such as authors, institutions, field of study) rather than the internal argumentative structure of papers.

Argumentation mining \cite{lawrence-reed-2019-argument} seeks to identify claim-premise relationships in text, but existing approaches focus on pre-defined structures, such as internet forums or student essays. Our work therefore extends these ideas to scientific papers, introducing a domain-specific ontology of 10 semantic roles and enforcing structural constraints like acyclicity and role-specific parent-child relationships.

Prior work on argumentative structure in scientific writing is also directly relevant here as existing studies analyze such structures in scientific articles \cite{kirschner-etal-2015-linking}, introduce argument-annotated corpora for scientific publications \cite{lauscher-etal-2018-argument}, and survey argument mining for processing academic documents \cite{al-khatib-etal-2021-argument}. These lines of work motivate our role-aware document representation, with Plato’s Cave differing by coupling document-structure extraction to external verification, trust propagation, and a reviewer-facing score rather than treating it as a standalone parsing task.

\subsection{Trust Propagation Methods}

In probabilistic graphical models, belief propagation (BP) is a canonical message-passing algorithm for propagating local evidence through a graph to compute marginal beliefs, with the sum-product algorithm providing a unifying view on factor graphs \cite{pearl1988probabilistic, kschischang2001factor}.
Loopy and generalized variants connect BP fixed points to variational objectives (e.g., Bethe-style free energies), providing a principled lens on approximate propagation in graphs with complex dependencies \cite{yedidia2003understanding, wainwright2008variational}.

Trust propagation has been widely studied in social network analysis \cite{10.1145/988672.988727} and citation networks \cite{334, 10.1145/324133.324140}. These approaches model trust as flowing through edges weighted by structural properties (citation count, user ratings). However, they lack grounding in content-based verification of the actual claims being made, which we incorporate to improve automated claim validation.

Recent work on detecting misinformation \cite{9620068} has incorporated source credibility and cross-source consistency, but these methods aggregate signals at the document level rather than modeling fine-grained dependency chains within the argumentation of a single paper. Our trust-gating mechanism differs by explicitly modeling how parent node quality gates the confidence of child nodes, preventing weak evidence from inflating downstream claims through a differentiable threshold function.

\subsection{Multi-Agent Systems}

Autonomous LLM-based agents with web-browsing capabilities have emerged as a powerful paradigm for information gathering \cite{DBLP:journals/corr/abs-2112-09332, browseruse}. Recent systems combine large language models with browser automation to answer questions, compare products, and fact-check claims \cite{magenticui}. Our work applies this approach to scientific verification by coordinating multiple agents to extract verification metrics across six dimensions (credibility, relevance, evidence strength, method rigor, reproducibility, and citation support).

\subsection{Positioning this Work in the Broader Literature}

We uniquely combine knowledge graph construction with content-based multi-agent verification and trust propagation in a learnable framework, aiming to expand fact checking from sentence-level judgments to complete argumentative structures. Unlike citation-network methods, our signals are grounded in verified content rather than popularity alone. This paper should therefore be read as a study on the proposed system and calibration schemes rather than as a definitive benchmark comparison against external baselines.

\section{Problem Formulation}\label{sec:problemformulation}

\begin{figure*}[t]
\centering
\includegraphics[width=\linewidth]{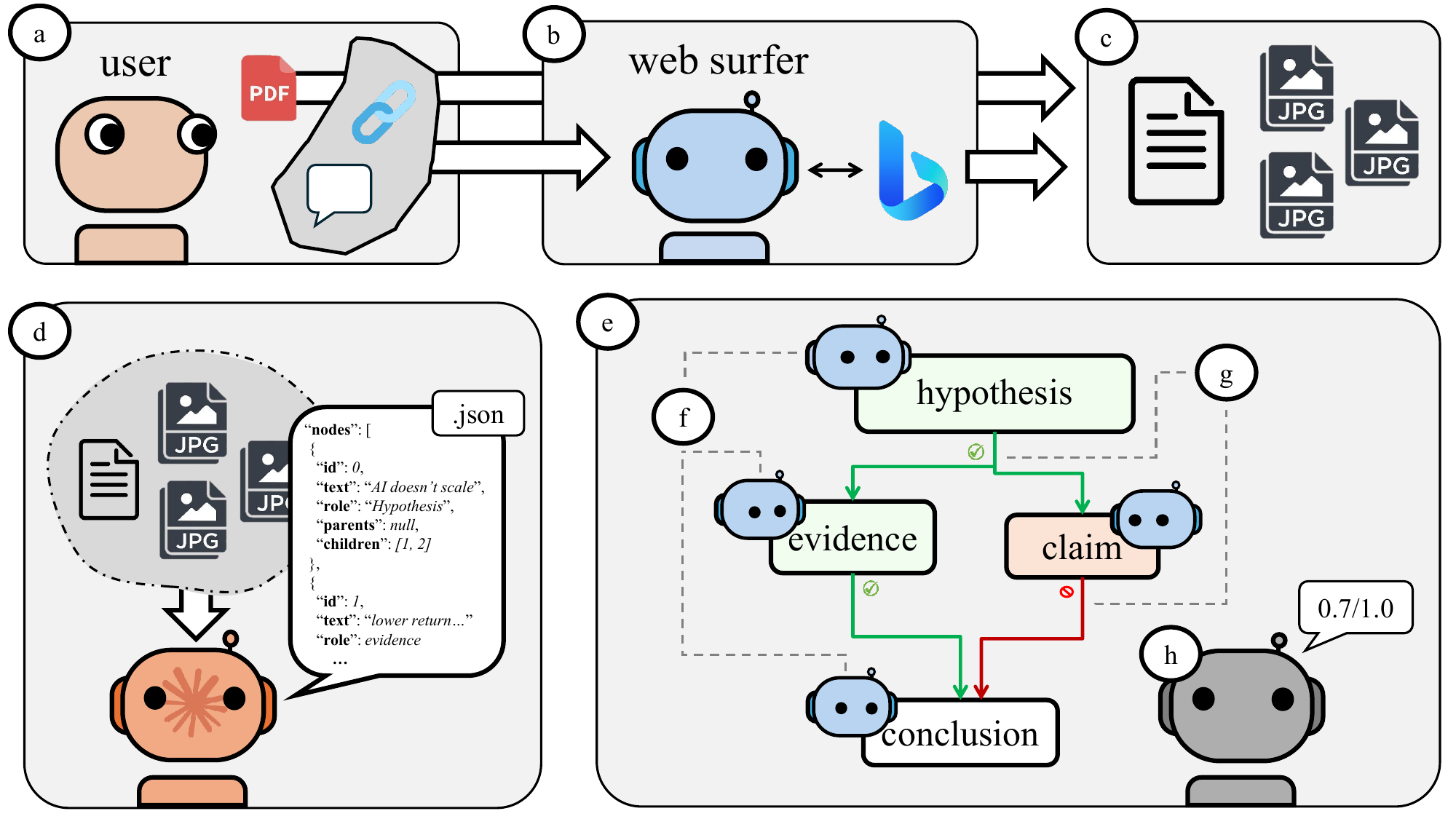}
\caption{System Overview. \textbf{(a)} The user provides a PDF, URL, or natural-language query. \textbf{(b)} For URLs or queries, a web-surfer agent browses for the specific paper. \textbf{(c)} All text and images from the document are extracted and stored. \textbf{(d)} The content is passed into an LLM and converted into a role-labeled DAG serialized as JSON (nodes and directed dependencies). \textbf{(e)} The frontend parses the DAG JSON and renders an interactive graph. \textbf{(f)} Web-surfer agents verify each node sequentially using external sources to produce normalized verification metrics. \textbf{(g)} Node qualities influence downstream nodes via trust-gated propagation along dependency edges. \textbf{(h)} The scorer aggregates node and edge signals into an overall paper-level score.}
\label{fig:system-overview}
\end{figure*}

We formalize the scoring problem as follows. Given a candidate knowledge graph extracted from a paper (nodes with roles and text, and directed dependency edges) together with per-node verification metrics, our goal is to compute calibrated node qualities, edge confidences, and an overall graph score. Formally, we represent the paper as a knowledge graph $\mathcal{G} = (V, E)$ where:
\begin{itemize}
    \item Each node $v \in V$ represents a semantic unit with role $\rho_v \in \mathcal{R}$ and quality score $q_v \in [0,1]$.
    \item Each edge $(u,v) \in E$ represents a logical dependency with confidence $C_{u \to v} \in [0,1]$.
    \item The graph $\mathcal{G}$ is a directed acyclic graph (DAG) respecting role-specific constraints.
\end{itemize}

The semantic role set is defined as $\mathcal{R}:= $ \{ Assumption, Claim, Conclusion, Context, Counterevidence, Evidence, Hypothesis, Limitation, Method, Result\} following the scorer's canonical role ontology. From $\mathcal{G}$, we compute an overall graph score $S_{\text{graph}}\in[0,1]$ that aggregates six interpretable components $\mathcal{T}:=$ \{ bridge coverage, best-path reliability, redundancy, fragility, coherence, and coverage \}. Our objective is for $S_{\text{graph}}$ to correlate with coarse human triage judgments while remaining decomposable into auditable sub-scores.

The coefficients $\{\gamma_t\}_{t\in\mathcal{T}}$ are calibration weights. Positive values reward desirable structural properties, while negative values encode penalties (e.g. fragility). The raw weighted sum is normalized back to $[0,1]$ after aggregation.

\begin{equation}
S_{\text{graph}} = \phi_{\mathrm{clip}[0,1]}\!\left(\sum_{t\in \mathcal{T}} \gamma_t\,S_t\right).
\end{equation}

The key challenges are addressed with the following \textbf{four-step} implementation. First, via \textbf{(1) Structure Extraction}, we parse unstructured text into a structured DAG with correct semantic roles. Then, through \textbf{(2) Claim verification}, we assess the validity of individual nodes using external knowledge. This is followed by \textbf{(3) Trust propagation} where node qualities flow through edges via a BP-inspired message-passing scheme such that weak evidence does not inflate downstream claims and finally \textbf{(4) Interpretability} ensures scores are explainable and aligned with human reasoning \cite{pearl1988probabilistic, kschischang2001factor, yedidia2003understanding}.

\section{Implementation}\label{sec:implementation}

This section describes the reference implementation of Plato's Cave,
as seen in Figure~\ref{fig:system-overview}. This is intentionally \emph{factorized} into many sections, including structure extraction, node verification, and graph scoring. These separate modules are connected by simple JSON artifacts, enabling independent iteration and ablation.

\begin{figure*}[t]
\centering
\fbox{\includegraphics[width=\linewidth]{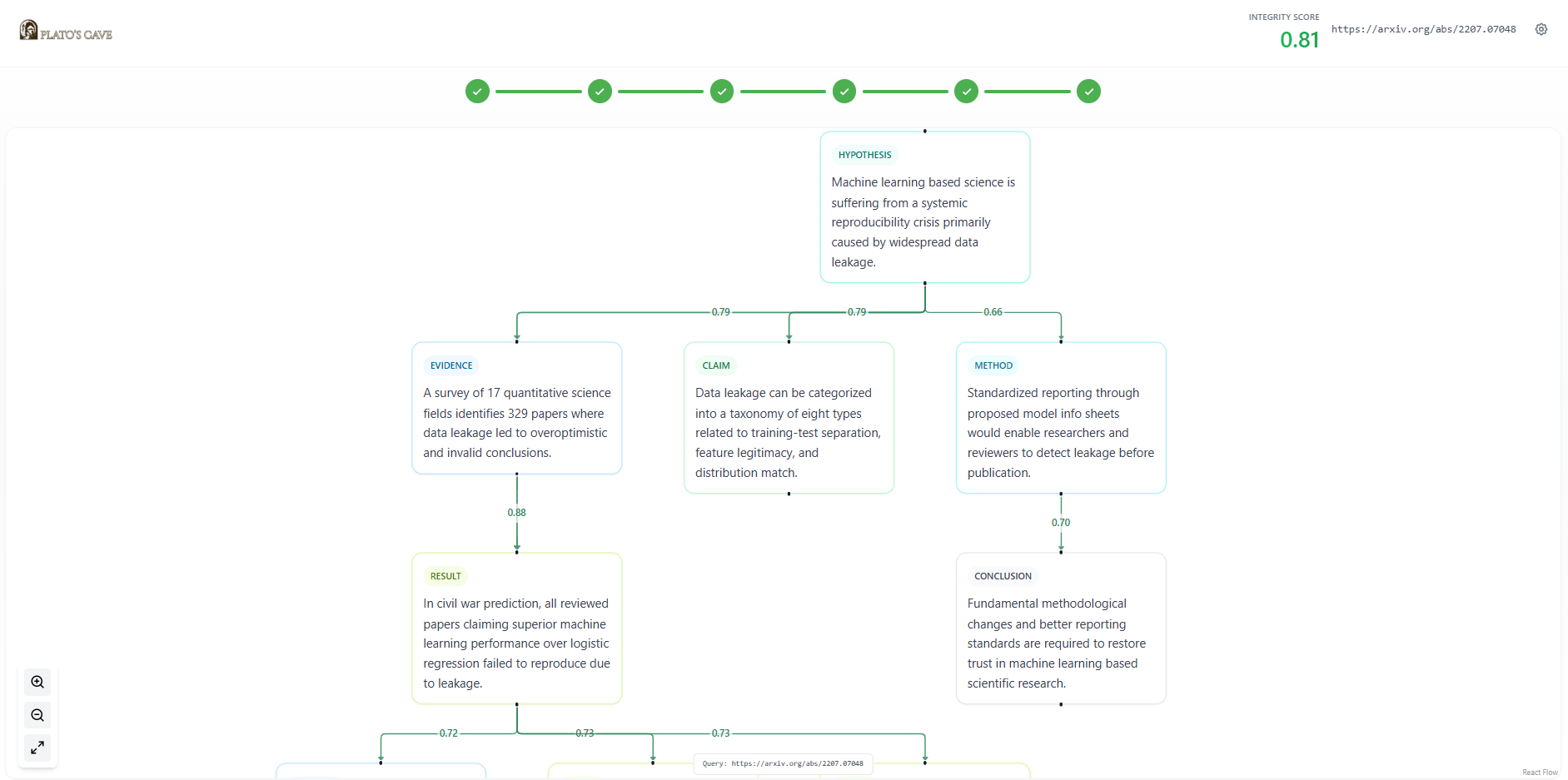}}
\caption{Plato's Cave interface showing a finalized run with the visualized DAG and Integrity Score}
\label{fig:system-screenshot}
\end{figure*}

\subsection{Knowledge Graph Construction}

We use a fixed ontology of 10 roles: Hypothesis, Claim, Evidence, Method, Result, Context, Assumption, Counterevidence, Limitation and Conclusion. Each node is a short excerpt of the paper (typically 1-3 sentences) labeled with one of these roles.
Edges represent \emph{dependency flow} from higher-level toward downstream synthesis; in the default priors, Hypothesis nodes are treated as roots and Conclusion nodes as sinks, with intermediate support roles (i.e. Evidence/Method/Result/Claim) acting as hypothesis to conclusion bridges.

A DAG extractor prompts an LLM with the paper text, the ontology, and a strict JSON schema (nodes with \texttt{id}, \texttt{role}, \texttt{text}, \texttt{parents}, \texttt{children}).
The output is validated for:
(i) ontology compliance, (i.e. the role of each node must be one of the 10 allowed values),
(ii) referential integrity (i.e. parent/child IDs must exist),
and (iii) acyclicity.
If validation fails, the system attempts a bounded ``repair'' pass by returning the validation errors to the LLM and requesting a corrected JSON. In Figure~\ref{fig:system-screenshot}, the DAG extraction uses \modelDAG\ and enforces a maximum of \maxNodesPerDAG\ nodes per sampled DAG.

\subsection{Node Verification}

Node verification is performed by an ensemble of LLM verifiers, where this agentic system is made available in a corresponding code release to aid replication. In this released scorer, verifier agents designate the presence or absence of six metrics per node (i.e. credibility, relevance, evidence strength, method rigor, reproducibility, citation support), each in $[0,1]$. Each verification agent is powered by browser-use, essentially an LLM with access to web search, navigation, and content extraction tools. The system also relies on Exa, an AI-powered search engine, as a fallback, in the event the web-surfer is unreachable. Agents receive a node (including the role and associated text snippet) and execute a search-verify-assess workflow:
\begin{enumerate}
    \item \textbf{Search}: Formulate search queries based on the node text and role (e.g., for Evidence nodes, search for the cited source; for Method nodes, search for validation studies).
    \item \textbf{Navigate}: Click search results, scroll to relevant sections, extract text, take screenshots when relevant.
    \item \textbf{Assess}: Evaluate source credibility (domain authority, publication venue), relevance (node claim alignment), and supporting evidence (data, citations).
\end{enumerate}
We use the browser-use library \cite{browseruse}, which optimizes LLM-browser interaction with action batching and caching, achieving 3-5x speedup over naive implementations. Agents run in Docker containers with remote browsers accessed via Chrome DevTools Protocol, enabling real-time visual monitoring through noVNC and interaction in the browser being used by the agent. For each node $v$, verification agents extract six metrics $\vecm_v = (m_{v,1}, \ldots, m_{v,6}) \in [0,1]^6$: \textbf{($m_1$) Credibility}, the trustworthiness of sources where peer-reviewed journals score higher than blogs); \textbf{($m_2$) Relevance}, the alignment between node claim and retrieved evidence; \textbf{ ($m_3$) Evidence strength}: the quality and quantity of supporting data; \textbf{($m_4$) Method Rigor}, the soundness of experimental design for method nodes; \textbf{($m_5$) Reproducibility}, the availability of code, data, or detailed procedures and 
\textbf{($m_6$) Citation support}, the number and quality of citations backing the claim.
The agent extracts these metrics by prompting the LLM with retrieved content and a structured output schema. Metrics are normalized and stored alongside the node.

\subsection{Trust-Propagating Graph Scoring}
\label{sec:trust_scoring}

This subsection summarizes the \emph{real-time} scoring backend used to evaluate a candidate paper-DAG after node-level metrics are available.
The goal is to assign (i) a \textbf{quality} to each node, (ii) a \textbf{confidence} to each dependency edge, and (iii) a \textbf{paper-level score} that reflects whether at least one hypothesis is supported by a coherent chain of evidence, method, and results.

For \textbf{inputs and update protocol} the scorer consumes a validated DAG $\mathcal{G}=(V,E)$ where each node $v\in V$ has:
(i) role $\rho_v \in \mathcal{R}$,
(ii) node text $t_v$,
and (iii) metrics $\mathbf{m}_v \in [0,1]^6$.
Node qualities and edge confidences are recomputed after metrics are available for a node and its parents, ensuring stability of intermediate results.

For the \textbf{node quality (role-aware metric fusion)} each node’s metrics are optionally reweighted by a \emph{global} per-metric weight vector, then fused into a single node quality score $q_v \in [0,1]$ using \emph{role-specific} weights.
Weights are normalized internally (by $\ell_1$ magnitude) and the resulting score is clipped to $[0,1]$.

For \textbf{edge confidence}, each directed edge $(u\!\to\!v)$ receives two related scores:

\begin{itemize}
    \item \textbf{Raw edge confidence} $C^{\text{raw}}_{u\to v}$: a local plausibility score computed as a weighted mixture of:
    (i) a role-transition prior for $(\rho_u,\rho_v)$,
    (ii) parent and child node qualities $(q_u,q_v)$,
    (iii) a lightweight lexical alignment score between node texts (Jaccard overlap), and
    (iv) a role-pair \emph{synergy} term that mixes parent and child metrics in a role-specific way.

    \item \textbf{Final edge confidence} $C_{u\to v}$: a trust-gated version of $C^{\text{raw}}_{u\to v}$ that down-weights dependencies coming from untrustworthy upstream nodes.
\end{itemize}

Following that, \textbf{trust propagation} assigns each node a propagated trust value $t_v \in [0,1]$ that combines its own quality $q_v$ with the ``weakest-link'' character of upstream support.
Each parent contributes an amount proportional to its trust (raised to an exponent $\alpha$) multiplied by the \emph{raw} edge confidence into $v$.
Parent contributions are aggregated using one of four modes: \texttt{min}, \texttt{mean}, \texttt{softmin}, or \texttt{dampmin}.
Finally, exposed edge confidence is gated by parent trust with a floor parameter $\eta$. Viewed through the lens of graphical-model inference, this is a deterministic message-passing rule analogous in spirit to (loopy) belief propagation / sum-product, but operating on bounded trust scores and incorporating an explicit trust gate to prevent unreliable parents from boosting descendants. \cite{kschischang2001factor, yedidia2003understanding, wainwright2008variational}

The \textbf{graph paper-level score} aggregates six interpretable components computed on the \emph{bridge} subgraph that connects at least one \textbf{Hypothesis} to at least one \textbf{Conclusion}:
bridge coverage, best-path reliability, redundancy (max-flow), fragility (min-cut, penalized), coherence, and coverage of key roles.
We treat best-path reliability as the primary \emph{argument strength} signal by selecting a single hypothesis to conclusion chain maximizing product confidence and reporting its length-normalized (geometric-mean) confidence.

The \textbf{learnable parameter surface} implementation exposes an interpretable parameter surface for calibration and ablations: global metric weights, role-specific node-quality weights, edge-combination weights, role transition priors, role-pair synergy specifications, propagation hyperparameters, and graph-score weights. We do not claim these defaults are uniquely correct a priori; instead, they define an auditable basis set whose contribution can be isolated through cache-first ablations over fixed DAG and node-score artifacts.

The current release does not implement gradient-based learning from human labels, as it focuses on transparent decomposition, feature export, and reproducible calibration.

\subsection{Cache-First Calibration and Ablation}
To calibrate the exposed parameter surface we use offline studies that reuse cached factorized runs rather than issuing new LLM calls. This separates the question of which signals the pipeline extracts from the question of how those signals should be aggregated.

\begin{enumerate}
    \item \textbf{Dense exploration}: We first search a broad parameter region over metric weights, edge-combination weights, propagation hyperparameters, and graph-level aggregation weights.
    \item \textbf{Local refinement}: We then narrow the search around the best-performing dense configurations to improve ranking discrimination while preserving interpretability.
    \item \textbf{Sparse fine-tuning}: Finally, we apply sparse local perturbations around the best refined configuration to test whether small coordinate-wise changes still improve the tuned objective.
\end{enumerate}

In parallel, the release includes cache-first ablations over node metrics, node roles, edge features, propagation settings, and graph-head components. These experiments are intended to test sensitivity and reproducibility of the aggregation rule, not to claim that the present ontology or weight choices are uniquely optimal.

\subsection{Architecture and Implementation}

We implement Plato's Cave as an open-source full-stack web application with modular components\footnote{Available at \url{https://github.com/matheusmaldaner/PlatosCave}}.

\begin{figure}[t]
\centering
\includegraphics[width=\linewidth]{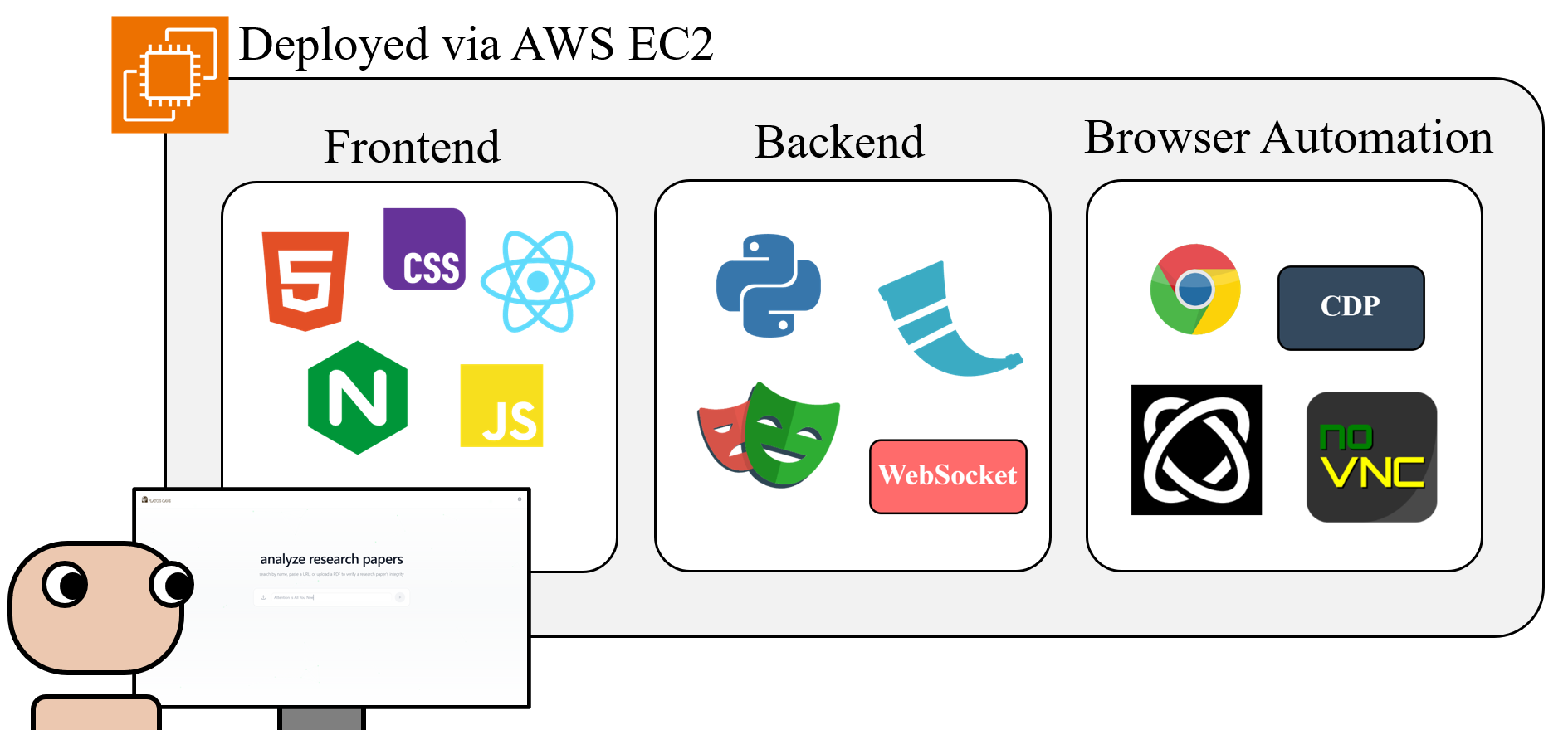}
\caption{System Architecture for Plato's Cave}
\label{fig:architecture}
\end{figure}

The frontend is built with Gatsby.js and React, providing an interactive graph visualization using ReactFlow with Dagre layout for hierarchical DAG rendering. Users upload PDFs, provide URLs or prompt the system with natural language, monitor real-time verification progress via WebSocket connections, and explore the resulting knowledge graph with hover tooltips displaying the six verification metrics per node and confidence breakdowns per edge. The interface highlights the best hypothesis-to-conclusion path and allows filtering by node role or quality threshold. Figure~\ref{fig:architecture} depicts the system architecture.

The release provides a self-contained scoring in Python:
\begin{itemize}
    \item \texttt{kg\_realtime\_scoring.py}: validation of KG JSON into a DAG, real-time updates to node qualities, trust values, edge confidences, and graph-level scoring; plus export utilities (node/edge feature matrices, random-walk corpus, and a paper fingerprint vector).
    \item \texttt{service\_adapter.py}: a thin session wrapper (\texttt{KGSession}) that (i) computes a BFS node order from Hypothesis roots, (ii) accepts metric updates for the current node, and (iii) returns updated edges and scores for UI/agent loops.
\end{itemize}

\subsection{Browser Automation}

We deploy browser-use agents in Docker containers with remote browsers accessed via Chrome DevTools Protocol. This architecture provides isolation (each agent runs in its own container), visual monitoring (noVNC allows observing agent actions in real-time), and scalability (horizontal scaling by launching more containers). The browser-use library optimizes LLM-browser interaction through action batching, DOM caching, and smart scrolling, improving efficiency over naive selenium-based approaches.

\subsection{Analysis Modes}
\label{sec:analysis_modes}
Scientific and technical documents differ substantially in structure, evidentiary norms, and verification requirements. A clinical trial, an arXiv preprint, and an SEC 10-k expose different failure modes and demand different extraction priorities. To address this heterogeneity without fragmenting the core pipeline, we introduce \emph{analysis modes}: lightweight, domain-specific configurations that modify claim extraction and graph construction while preserving a shared verification and scoring backend. The system additionally supports three mutually exclusive modes, selectable via a CLI flag (\texttt{--mode}) or frontend user interface:
\begin{itemize}
    \item \textbf{Academic}: preprints, theses, conference papers; emphasizes argument structure and theoretical grounding.
    \item \textbf{Journal}: peer-reviewed articles and clinical studies; emphasizes statistical rigor, replication, and disclosure.
    \item \textbf{Finance}: earnings reports, SEC filings, investor materials; emphasizes financial metrics and risk disclosures.
\end{itemize}
Modes act exclusively at DAG extraction time by injecting domain-specific instructions into the LLM prompt. All modes share the same base ontology of 10 universal roles (i.e. Hypothesis, Claim, Evidence, Method, Result, Conclusion, Assumption, Counterevidence, Limitation, Context).



The extractor prompt specifies the expanded role set and mode-specific prioritization rules (e.g., extracting exact $p$-values and confidence intervals in Journal mode, or separating historical data from forward-looking statements in Finance mode).
DAG validation accepts the union of base and mode-specific roles, ensuring schema consistency across modes.

Importantly, modes do \emph{not} alter downstream verification or scoring logic.
Once a valid DAG is produced, node verification agents and trust-propagating graph scoring operate identically across modes.
This design choice isolates domain assumptions to extraction only, enabling controlled ablations and preventing silent changes to scoring semantics.


\subsection{Design Rationale}
Several architectural decisions merit explanation. We chose a multi-agent architecture over a single sequential agent because verification tasks are independent and parallelization reduces total latency from minutes to seconds. We separated the graph scorer into its own service to enable offline experimentation with scoring parameters without re-running verification. We use WebSockets for real-time updates rather than polling to provide responsive user feedback during the lengthy verification process. We enforce DAG constraints at generation time rather than post-hoc to prevent invalid graphs from reaching verification, saving compute resources.

The modular design allows future extensions such as replacing the LLM backend, swapping verification agents (e.g., defining custom functions for tool calling), and integrating learned parameters (by exposing the scorer's gradient interface) with Figure~\ref{fig:architecture} depicting our implementation.

\section{Evaluation}\label{sec:evaluation}

We evaluate Plato's Cave as an \emph{end-to-end verification pipeline} and as a \emph{scoring system}. The attached results correspond to a pilot run (\texttt{runs/experiments\_debug4}) intended to validate throughput, artifact quality, and failure modes prior to a larger human-annotated evaluation.

\subsection{Experimental Setup}

The run loaded $\numPapersProcessed$ paper records from a spreadsheet collection spanning three topical sheets (Economics, ML/Computing, and Psychology). 
Each processed paper also has a spreadsheet triage label in $\{\texttt{Bad}, \texttt{Neutral}, \texttt{Good}\}$ assigned during collection curation. We use these labels only as weak supervision for calibration: they summarize an overall document-level judgment of the paper's central claims and practical credibility, rather than serving as gold labels for every extracted node or edge. Accordingly, we treat this evaluation as a noisy ranking signal rather than a definitive quality benchmark.
For each processed paper we sample $K=\numDAGSamplesPerPaper$ candidate DAGs and, for each \emph{valid} DAG, run $M=\numNodeWeightSamples$ scoring trials by resampling node-quality/propagation weights.
This produced $\numDAGSamplesValid$ valid DAG samples and $\numTrialsTotal$ total trials.
DAG extraction uses \modelDAG; node verification uses \modelNode.
The run uses bounded concurrency at three levels (papers, nodes, and global LLM calls) to avoid rate-limit collapse while maintaining parallelism.

Regarding \textbf{Throughput}, the full collection run completed in $\wallclockHours$ hours wall-clock time.
Among successfully processed papers, the median per-paper runtime was $\paperRuntimeMedianMin$ minutes (90th percentile $\textbf{70.7}$ minutes). 

\subsection{DAG Validity and Repair}

Across $\numDAGSamplesTotal$ sampled DAGs, $\numDAGSamplesInvalid$ failed validation (2.6\%).
The most common failure modes were (i) \emph{unknown role labels} (e.g., ``Recommendation'', ``Example'') and (ii) \emph{dangling references} where a parent/child ID did not exist.
These failures are actionable: tightening the extraction prompt to forbid out-of-ontology role names and to require explicit, consistent IDs reduces wasted trials and improves effective throughput.

\subsection{Ranking Performance and Calibration}

We evaluate each paper's mean score (averaged over all valid DAG samples and node-weight trials) against the spreadsheet's coarse $\texttt{Bad}$ / $\texttt{Neutral}$ / $\texttt{Good}$ label. Because these labels are weak supervision, we treat this as a calibration problem rather than a definitive benchmark and our experiments emphasize cache-first staged search and internal ablations over fixed cached artifacts. We additionally do not provide any external baseline comparison.

The calibration study uses three consecutive search stages over the cached \texttt{runs/experiments\_debug4} artifacts: dense, refine, and sparse. For binary Good-vs-Bad discrimination, we report AUROC (area under the receiver operating characteristic curve), where 0.5 indicates chance-level separation and 1.0 indicates perfect separation. On the 104 papers whose labels are exactly $\texttt{Good}$, $\texttt{Neutral}$, or $\texttt{Bad}$, the best configurations achieve:

\begin{center}
\begin{tabular}{l c c}
\toprule
Search Stage & Best AUROC & Best Spearman \\
\midrule
Dense search & 0.7025 & 0.313 \\
Refine search & 0.7594 & 0.395 \\
Sparse stage-3 & 0.7658 & 0.401 \\
\bottomrule
\end{tabular}
\end{center}

\begin{figure}[t]
\centering
\includegraphics[width=\linewidth]{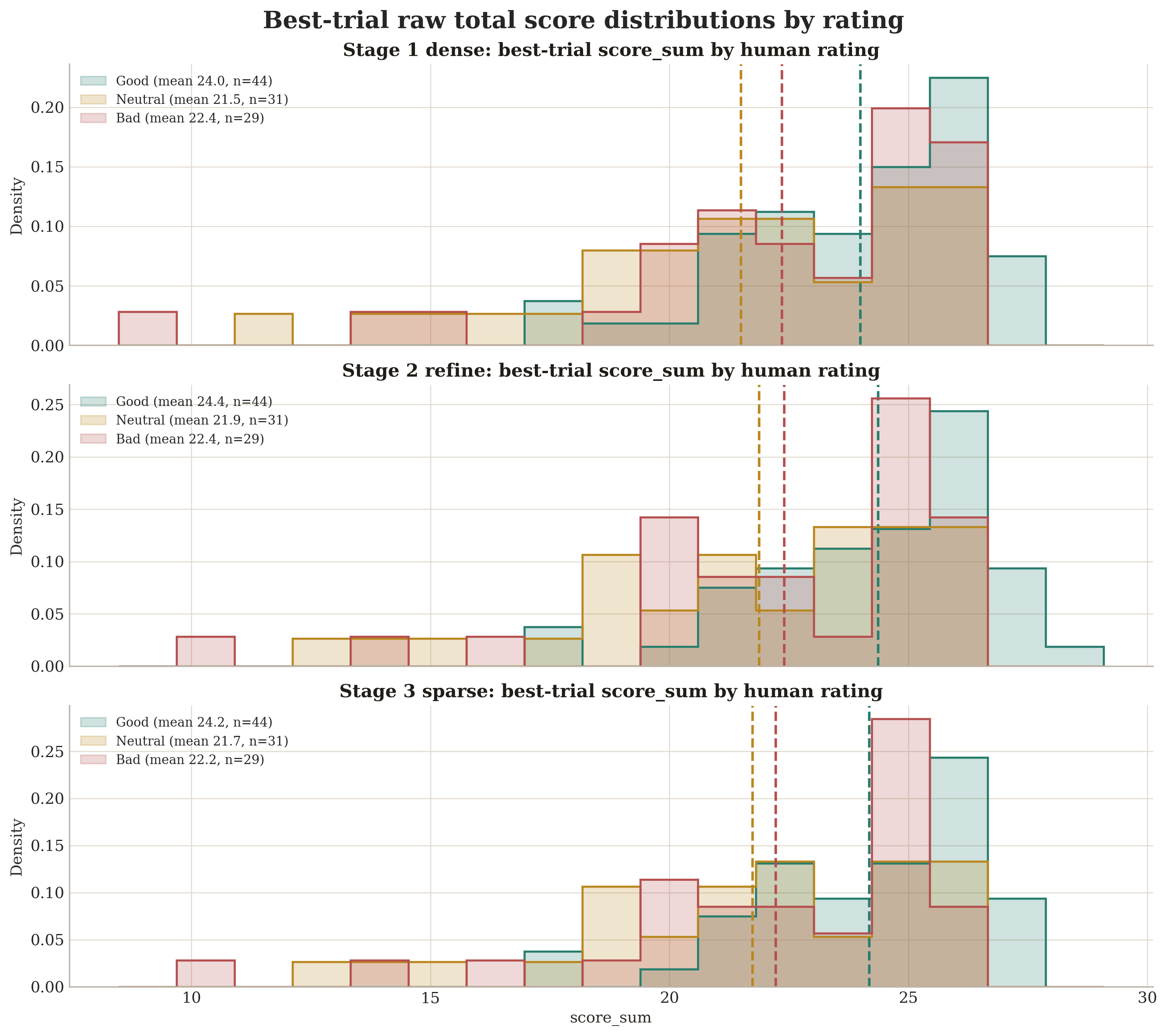}
\caption{Paper-level mean score grouped by spreadsheet rating under the selected calibrated setting. We use a discrete, non-smoothed summary to avoid overstating distributional structure under weak supervision.}
\label{fig:eval_score_rating}
\end{figure}

Performance improves monotonically across the tuned objective, the final sparse stage yields a narrow gain over the refine stage, showing signs of overfitting on this benchmark. For that reason, we treat the refine-stage configuration as the main focus and the sparse stage as evidence that the scoring surface requires further optimization.

\subsection{Interpretation}

Relative to the earlier untuned scorer, offline calibration produces moderate Good-vs-Bad separation and moderate ordinal agreement with the coarse triage labels. At the same time, the weakness of the supervision signal and the small stage-3 gain argue against stronger claims: the current evidence supports Plato's Cave as a tunable, auditable ranking scaffold, not yet as a definitive paper-quality estimator.

From the performance log, the median per-DAG extraction time was \textbf{43.1}\,s and the median per ``score-nodes-once'' trial time was \textbf{53.6}\,s (measured on the subset of papers captured in the perf log).
These latencies imply that DAG extraction dominates per-paper cost under the $K\times M$ sampling regime.

\section{Discussion and Future Work}\label{sec:discussion_futurework}

While the current system demonstrates that trust-propagating graph scoring over LLM-extracted DAGs can produce moderate alignment with human triage labels, there are many directions to be explored. This section outlines the strengths, limitations and future areas of work.

\subsection{Strengths}

Our trust-propagating knowledge graph approach offers several advantages over existing methods. The system provides interpretability by exposing the complete reasoning chain from evidence to conclusions, allowing users to inspect which nodes are weak and why. The trust-gating mechanism addresses a fundamental limitation of citation networks and knowledge graphs: it prevents weak evidence from inflating strong claims by modeling parent quality as a gate function. The multi-agent verification architecture grounds assessments in external knowledge rather than relying solely on parametric knowledge in language models. The differentiable scoring framework enables learning from human feedback, allowing continuous improvement with more annotations.

The attached release includes staged tuning over fixed artifacts, which make the scorer reproducible and inspectable. However, these studies should be interpreted as calibration evidence for the aggregation rule, not as proof that the current ontology, equations, or default weights are uniquely correct.

\subsection{Limitations}

Several limitations warrant discussion. The system requires substantial computational resources, particularly for multi-agent verification which involves multiple LLM API calls and browser automation sessions per paper. In the full collection run, the median end-to-end per-paper runtime was \textbf{$\paperRuntimeMedianMin$ minutes} and costs approximately \textbf{ \$ $\costAPI$ } in API fees. This limits scalability to real-time monitoring of preprint servers.

Our evaluation dataset contains \textbf{104} papers, which is modest compared to large-scale fact-checking benchmarks. We focus on depth (fine-grained annotations, multiple metrics) over breadth, but larger-scale evaluation would strengthen our claims. Future work should explore multi-annotator aggregation methods.

The semantic role ontology with 10 roles captures many aspects of scientific argumentation but may require domain-specific extensions. For example, biology papers might need roles for ``Organism Model'' or ``Experimental Control,'' while computer science papers might need ``Proof Sketch'' nodes. We currently use a fixed ontology across domains, which may limit performance on specialized fields.

The trust-gate and aggregation parameters are currently selected by offline cache-first search rather than learned from human annotations. This is an improvement over purely hand-set defaults, but it still does not establish that the present weights will generalize beyond this benchmark or beyond the current weak supervision signal.

The system is human-centered because the reviewer and verification agent share the same live browser session. Reviewers can see which pages are being visited in real time and intervene when needed (for example, solving CAPTCHAs or entering university affiliation credentials to access paywalled sources), then allow the agent to continue verification with that access.

\subsection{Future Work}

We identify several promising research directions. Learning the trust gate parameters and role-specific weights from human annotations could improve performance and domain adaptation. Active learning could identify which nodes to verify next, reducing verification costs while maintaining accuracy. Cross-domain generalization experiments testing whether a model trained on computer science papers transfers to biology would assess the robustness of our semantic ontology. Integration with peer review systems could provide real-time feedback to reviewers and authors during the submission process.

Longitudinal monitoring of preprint servers like arXiv could flag potentially problematic papers early, before formal peer review; recent work on agents that can wait, monitor, and act over extended periods \cite{sentinelstep2025} suggests that such sustained surveillance is becoming practical. Extending the system to handle multi-paper claims (e.g., a claim supported by evidence from multiple papers) would enable analysis of research programs rather than individual papers. Formalizing the types of explanations our system produces (trust scores, graph structure, verification chains) using general XAI syntax \cite{oconnor2025xaisyntax} could standardize the interface and guide the design of richer, neurosymbolic explanation strategies. Explanations in natural language would further make the system more accessible to non-expert users.

Finally, exploring alternative verification methods beyond web search (e.g., running code to reproduce results, querying structured databases like PubMed) could improve accuracy for claims amenable to automated checking.

\section{Conclusion}\label{sec:conclusion}

We presented Plato's Cave, a human-centered system for scientific paper verification that combines role-aware knowledge graphs, external node verification, and trust-gated score propagation. This produces interpretable DAGs, node/edge metrics, and auditable score traces, for reviewer inspection.

On a heterogeneous collection of 104 papers the factorized pipeline was operationally stable: only \textbf{2.6\%} of DAG samples failed strict validation, and \textbf{97.4\%} of attempted scoring trials completed successfully. Cache-first calibration then achieved Good-vs-Bad AUROC \textbf{0.759} at the conservative refine stage (\textbf{0.766} in the final sparse stage) with Spearman \textbf{0.395}/\textbf{0.401} on the 104 papers with exact \texttt{Good}/\texttt{Neutral}/\texttt{Bad} labels.

Overall, the current evidence supports Plato's Cave as a reproducible, auditable decision-support scaffold rather than a definitive paper-quality estimator. The main next steps are stronger supervision, external baselines, and explanation interfaces tied to specific low-trust nodes and edges.

\bibliography{aaai2026} 

@techreport{NSB2023,
  author = {{National Science Board}},
  title = {Publication Output by Region, Country, or Economy and by Scientific Field},
  booktitle = {Publications Output: U.S. Trends and International Comparisons},
  institution = {National Science Foundation},
  year = {2023},
  type = {NSB-2023-33},
  address = {Alexandria, VA},
  month = dec,
  url = {https://ncses.nsf.gov/pubs/nsb202333/publication-output-by-region-country-or-economy-and-by-scientific-field}
}

@misc{RetractionWatch2024,
  author = {{Retraction Watch}},
  year = 2024,
  title = "{Retraction Watch Database}",
  howpublished = "\url{https://retractionwatch.com/2024/}",
  note = "Accessed: 2026-01-01"
}

@misc{RetractionWatch2025,
  author = {{Retraction Watch}},
  year = 2025,
  title = "{Cheers, 2025: Retraction Watch Turned 15, Center for Scientific Integrity}",
  howpublished = "\url{https://retractionwatch.com/2025/12/30/cheers-2025-retraction-watch-turned-15-center-for-scientific-integrity/}",
  note = "Accessed: 2026-01-01"
}

@article{strathern:97,
  author  = "Strathern, Marilyn",
  title   = "{'Improving Ratings': Audit in the British University System}",
  journal = "European Review",
  volume  = 5,
  number  = 3,
  pages   = "305--321",
  year    = 1997,
  doi     = "10.1017/S1062798700002660"
}

@misc{wadden2020factfictionverifyingscientific,
      title={Fact or Fiction: Verifying Scientific Claims}, 
      author={David Wadden and Shanchuan Lin and Kyle Lo and Lucy Lu Wang and Madeleine van Zuylen and Arman Cohan and Hannaneh Hajishirzi},
      year={2020},
      eprint={2004.14974},
      archivePrefix={arXiv},
      primaryClass={cs.CL},
      url={https://arxiv.org/abs/2004.14974}, 
}

@misc{thorne2018feverlargescaledatasetfact,
      title={FEVER: a large-scale dataset for Fact Extraction and VERification}, 
      author={James Thorne and Andreas Vlachos and Christos Christodoulopoulos and Arpit Mittal},
      year={2018},
      eprint={1803.05355},
      archivePrefix={arXiv},
      primaryClass={cs.CL},
      url={https://arxiv.org/abs/1803.05355}, 
}

@article{liu2023reviewergpt,
  author  = "Liu, Ryan and Shah, Nihar B.",
  title   = "{ReviewerGPT? An Exploratory Study on Using Large Language Models for Paper Reviewing}",
  journal = "arXiv preprint arXiv:2306.00622",
  year    = 2023
}

@inbook{Liu_Datta_Lim_2015, 
place={Boca Raton, Florida}, 
title={Judging the Veracity of Claims and Reliability of Sources }, booktitle={Computational Trust Models and Machine Learning}, 
publisher={CRC Press},
author={Liu, Xin and Datta, Anwitaman and Lim, Ee-Peng}, 
year={2015}, 
pages={40–71}}

@inproceedings{yang2018hotpotqa,
  title={{HotpotQA}: A Dataset for Diverse, Explainable Multi-hop Question Answering},
  author={Yang, Zhilin and Qi, Peng and Zhang, Saizheng and Bengio, Yoshua and Cohen, William W. and Salakhutdinov, Ruslan and Manning, Christopher D.},
  booktitle={Conference on Empirical Methods in Natural Language Processing ({EMNLP})},
  year={2018}
}

@article{Kinney2023TheSS,
  title={The Semantic Scholar Open Data Platform},
  author={Rodney Michael Kinney and Chloe Anastasiades and Russell Authur and Iz Beltagy and Jonathan Bragg and Alexandra Buraczynski and Isabel Cachola and Stefan Candra and Yoganand Chandrasekhar and Arman Cohan and Miles Crawford and Doug Downey and Jason Dunkelberger and Oren Etzioni and Rob Evans and Sergey Feldman and Joseph Gorney and David W. Graham and F.Q. Hu and Regan Huff and Daniel King and Sebastian Kohlmeier and Bailey Kuehl and Michael Langan and Daniel Lin and Haokun Liu and Kyle Lo and Jaron Lochner and Kelsey MacMillan and Tyler C. Murray and Christopher Newell and Smita Rao and Shaurya Rohatgi and Paul Sayre and Shannon Zejiang Shen and Amanpreet Singh and Luca Soldaini and Shivashankar Subramanian and A. Tanaka and Alex D Wade and Linda M. Wagner and Lucy Lu Wang and Christopher Wilhelm and Caroline Wu and Jiangjiang Yang and Angele Zamarron and Madeleine van Zuylen and Daniel S. Weld},
  journal={ArXiv},
  year={2023},
  volume={abs/2301.10140},
  url={https://api.semanticscholar.org/CorpusID:256194545}
}

@inproceedings{10.1145/3292500.3330785,
    author = {Zhang, Fanjin and Liu, Xiao and Tang, Jie and Dong, Yuxiao and Yao, Peiran and Zhang, Jie and Gu, Xiaotao and Wang, Yan and Shao, Bin and Li, Rui and Wang, Kuansan},
    title = {OAG: Toward Linking Large-scale Heterogeneous Entity Graphs},
    year = {2019},
    isbn = {9781450362016},
    publisher = {Association for Computing Machinery},
    address = {New York, NY, USA},
    url = {https://doi.org/10.1145/3292500.3330785},
    doi = {10.1145/3292500.3330785},
    booktitle = {Proceedings of the 25th ACM SIGKDD International Conference on Knowledge Discovery \& Data Mining},
    pages = {2585–2595},
    numpages = {11},
    location = {Anchorage, AK, USA},
    series = {KDD '19}
}

@inproceedings{sinha2015an,
    author = {Sinha, Arnab and Shen, Zhihong and Song, Yang and Ma, Hao and Eide, Darrin and Wang, Kuansan},
    title = {An Overview of Microsoft Academic Service (MAS) and Applications},
    booktitle = {WWW - World Wide Web Consortium (W3C)},
    year = {2015},
    month = {May},
    url = {https://www.microsoft.com/en-us/research/publication/an-overview-of-microsoft-academic-service-mas-and-applications-2/},
}

@inproceedings{10.1145/1401890.1402008,
author = {Tang, Jie and Zhang, Jing and Yao, Limin and Li, Juanzi and Zhang, Li and Su, Zhong},
title = {ArnetMiner: extraction and mining of academic social networks},
year = {2008},
isbn = {9781605581934},
publisher = {Association for Computing Machinery},
address = {New York, NY, USA},
url = {https://doi.org/10.1145/1401890.1402008},
doi = {10.1145/1401890.1402008},
booktitle = {Proceedings of the 14th ACM SIGKDD International Conference on Knowledge Discovery and Data Mining},
pages = {990–998},
numpages = {9},
location = {Las Vegas, Nevada, USA},
series = {KDD '08}
}

@article{lawrence-reed-2019-argument,
    title = "Argument Mining: A Survey",
    author = "Lawrence, John  and
      Reed, Chris",
    journal = "Computational Linguistics",
    volume = "45",
    number = "4",
    month = dec,
    year = "2019",
    address = "Cambridge, MA",
    publisher = "MIT Press",
    url = "https://aclanthology.org/J19-4006/",
    doi = "10.1162/coli_a_00364",
    pages = "765--818"
}

@inproceedings{10.1145/988672.988727,
    author = {Guha, R. and Kumar, Ravi and Raghavan, Prabhakar and Tomkins, Andrew},
    title = {Propagation of trust and distrust},
    year = {2004},
    isbn = {158113844X},
    publisher = {Association for Computing Machinery},
    address = {New York, NY, USA},
    url = {https://doi.org/10.1145/988672.988727},
    doi = {10.1145/988672.988727},
    booktitle = {Proceedings of the 13th International Conference on World Wide Web},
    pages = {403–412},
    numpages = {10},
    location = {New York, NY, USA},
    series = {WWW '04}
}

@article{334,
    title	= {The Anatomy of a Large-Scale Hypertextual Web Search Engine},
    author	= {Sergey Brin and Lawrence Page},
    year	= {1998},
    URL	= {https://snap.stanford.edu/class/cs224w-readings/Brin98Anatomy.pdf},
    journal	= {Computer Networks},
    pages	= {107-117},
    volume	= {30}
}

@article{10.1145/324133.324140,
    author = {Kleinberg, Jon M.},
    title = {Authoritative sources in a hyperlinked environment},
    year = {1999},
    issue_date = {Sept. 1999},
    publisher = {Association for Computing Machinery},
    address = {New York, NY, USA},
    volume = {46},
    number = {5},
    issn = {0004-5411},
    url = {https://doi.org/10.1145/324133.324140},
    doi = {10.1145/324133.324140},
    journal = {J. ACM},
    month = sep,
    pages = {604–632},
    numpages = {29}
}

@ARTICLE{9620068,
  author={Mridha, M. F. and Keya, Ashfia Jannat and Hamid, Md. Abdul and Monowar, Muhammad Mostafa and Rahman, Md. Saifur},
  journal={IEEE Access}, 
  title={A Comprehensive Review on Fake News Detection With Deep Learning}, 
  year={2021},
  volume={9},
  number={},
  pages={156151-156170},
  doi={10.1109/ACCESS.2021.3129329}
}

@article{DBLP:journals/corr/abs-2112-09332,
  author       = {Reiichiro Nakano and
                  Jacob Hilton and
                  Suchir Balaji and
                  Jeff Wu and
                  Long Ouyang and
                  Christina Kim and
                  Christopher Hesse and
                  Shantanu Jain and
                  Vineet Kosaraju and
                  William Saunders and
                  Xu Jiang and
                  Karl Cobbe and
                  Tyna Eloundou and
                  Gretchen Krueger and
                  Kevin Button and
                  Matthew Knight and
                  Benjamin Chess and
                  John Schulman},
  title        = {WebGPT: Browser-assisted question-answering with human feedback},
  journal      = {CoRR},
  volume       = {abs/2112.09332},
  year         = {2021},
  url          = {https://arxiv.org/abs/2112.09332},
  eprinttype    = {arXiv},
  eprint       = {2112.09332},
  bibsource    = {dblp computer science bibliography, https://dblp.org}
}

@misc{browseruse,
    author={Browser-Use},
    year={2024},
    howpublished="\url{https://browser-use.com/}",
    note="Accessed: 2026-01-03"
}

@Article{Liu2020,
    author={Liu, Kang},
    title={A survey on neural relation extraction},
    journal={Science China Technological Sciences},
    year={2020},
    month={Oct},
    day={01},
    volume={63},
    number={10},
    pages={1971-1989},
    issn={1869-1900},
    doi={10.1007/s11431-020-1673-6},
    url={https://doi.org/10.1007/s11431-020-1673-6}
}

@misc{zhu2024llmsknowledgegraphconstruction,
      title={LLMs for Knowledge Graph Construction and Reasoning: Recent Capabilities and Future Opportunities}, 
      author={Yuqi Zhu and Xiaohan Wang and Jing Chen and Shuofei Qiao and Yixin Ou and Yunzhi Yao and Shumin Deng and Huajun Chen and Ningyu Zhang},
      year={2024},
      eprint={2305.13168},
      archivePrefix={arXiv},
      primaryClass={cs.CL},
      url={https://arxiv.org/abs/2305.13168}, 
}

@article{CHENG202495,
    title = {Automated knowledge graphs for complex systems (AutoGraCS): Applications to management of bridge networks},
    journal = {Resilient Cities and Structures},
    volume = {3},
    number = {4},
    pages = {95-106},
    year = {2024},
    issn = {2772-7416},
    doi = {https://doi.org/10.1016/j.rcns.2024.11.001},
    url = {https://www.sciencedirect.com/science/article/pii/S2772741624000607},
    author = {Minghui Cheng and Syed M.H. Shah and Antonio Nanni and H. Oliver Gao}
    }

@book{pearl1988probabilistic,
  author    = {Pearl, Judea},
  title     = {Probabilistic Reasoning in Intelligent Systems: Networks of Plausible Inference},
  publisher = {Morgan Kaufmann Publishers},
  address   = {San Mateo, CA, USA},
  year      = {1988},
  isbn      = {1558604790}
}

@article{kschischang2001factor,
  author  = {Kschischang, Frank R. and Frey, Brendan J. and Loeliger, Hans{-}Andrea},
  title   = {Factor Graphs and the Sum-Product Algorithm},
  journal = {IEEE Transactions on Information Theory},
  volume  = {47},
  number  = {2},
  pages   = {498--519},
  year    = {2001},
  month   = feb,
  doi     = {10.1109/18.910572}
}

@incollection{yedidia2003understanding,
  author    = {Yedidia, Jonathan S. and Freeman, William T. and Weiss, Yair},
  title     = {Understanding Belief Propagation and its Generalizations},
  booktitle = {Exploring Artificial Intelligence in the New Millennium},
  editor    = {Lakemeyer, Gerhard and Nebel, Bernhard},
  publisher = {Morgan Kaufmann Publishers Inc.},
  address   = {San Francisco, CA, USA},
  pages     = {239--269},
  year      = {2003},
  doi       = {10.5555/779343.779352}
}

@article{wainwright2008variational,
  author  = {Wainwright, Martin J. and Jordan, Michael I.},
  title   = {Graphical Models, Exponential Families, and Variational Inference},
  journal = {Foundations and Trends in Machine Learning},
  volume  = {1},
  number  = {1-2},
  pages   = {1--305},
  year    = {2008},
  doi     = {10.1561/2200000001}
}

@inproceedings{kirschner-etal-2015-linking,
    title = "Linking the Thoughts: Analysis of Argumentation Structures in Scientific Publications",
    author = "Kirschner, Christian and
      Eckle-Kohler, Judith and
      Gurevych, Iryna",
    booktitle = "Proceedings of the 2nd Workshop on Argumentation Mining",
    month = jun,
    year = "2015",
    address = "Denver, CO",
    publisher = "Association for Computational Linguistics",
    url = "https://aclanthology.org/W15-0501/",
    doi = "10.3115/v1/W15-0501",
    pages = "1--11"
}

@inproceedings{lauscher-etal-2018-argument,
    title = "An Argument-Annotated Corpus of Scientific Publications",
    author = "Lauscher, Anne and
      Glava{\v{s}}, Goran and
      Ponzetto, Simone Paolo",
    booktitle = "Proceedings of the 5th Workshop on Argument Mining",
    month = nov,
    year = "2018",
    address = "Brussels, Belgium",
    publisher = "Association for Computational Linguistics",
    url = "https://aclanthology.org/W18-5206/",
    doi = "10.18653/v1/W18-5206",
    pages = "40--46"
}

@inproceedings{al-khatib-etal-2021-argument,
    title = "Argument Mining for Scholarly Document Processing: Taking Stock and Looking Ahead",
    author = "Al Khatib, Khalid and
      Ghosal, Tirthankar and
      Hou, Yufang and
      de Waard, Anita and
      Freitag, Dayne",
    booktitle = "Proceedings of the Second Workshop on Scholarly Document Processing",
    month = jun,
    year = "2021",
    address = "Online",
    publisher = "Association for Computational Linguistics",
    url = "https://aclanthology.org/2021.sdp-1.7/",
    doi = "10.18653/v1/2021.sdp-1.7",
    pages = "56--65"
}

@misc{sentinelstep2025,
  author = {Mozannar, Hussein and Maldaner, Matheus Kunzler and Murad, Maya and Chen, Jingya and Bansal, Gagan and Hosn, Rafah and Fourney, Adam},
  title = {Tell Me When: Building Agents That Can Wait, Monitor, and Act},
  howpublished = {\url{https://www.microsoft.com/en-us/research/blog/tell-me-when-building-agents-that-can-wait-monitor-and-act/}},
  year = {2025},
  note = {Microsoft Research Blog. Accessed: 2026-03-01}
}

@article{oconnor2025xaisyntax,
  author  = {Wormald, Stephen and Maldaner, Matheus Kunzler and O'Connor, Kristian D. and Dizon-Paradis, Olivia P. and Woodard, Damon L.},
  title   = {Abstracting general syntax for {XAI} after decomposing explanation sub-components},
  journal = {Artificial Intelligence Review},
  year    = {2025},
  doi     = {10.1007/s10462-025-11216-8}
}

@misc{neurips2025blog,
  author = {{NeurIPS 2025 Program Chairs}},
  title = {Reflections on the 2025 Review Process from the Program Committee Chairs},
  howpublished = {\url{https://blog.neurips.cc/2025/09/30/reflections-on-the-2025-review-process-from-the-program-committee-chairs/}},
  year = {2025},
  note = {Accessed: 2026-03-01}
}

@misc{papercopilot2026,
  author = {{Paper Copilot}},
  title = {Conference Statistics: {ICML}, {ICLR}, {NeurIPS}},
  howpublished = {\url{https://papercopilot.com/statistics/}},
  year = {2026},
  note = {Accessed: 2026-03-01}
}

@misc{magenticui,
      title={Magentic-UI: Towards Human-in-the-loop Agentic Systems}, 
      author={Hussein Mozannar and Gagan Bansal and Cheng Tan and Adam Fourney and Victor Dibia and Jingya Chen and Jack Gerrits and Tyler Payne and Matheus Kunzler Maldaner and Madeleine Grunde-McLaughlin and Eric Zhu and Griffin Bassman and Jacob Alber and Peter Chang and Ricky Loynd and Friederike Niedtner and Ece Kamar and Maya Murad and Rafah Hosn and Saleema Amershi},
      year={2025},
      eprint={2507.22358},
      archivePrefix={arXiv},
      primaryClass={cs.AI},
      url={https://arxiv.org/abs/2507.22358}, 
}

\appendix

\section{System Prompts}

\begin{promptbox}{Paper Extraction Prompt}
Your name is Plato, you are an expert academic paper finder and analyzer.
\medskip

Your mission: FIND the academic paper (if needed) and extract its CORE CONTENT. Work FAST --- scan and extract, don't read every word.
\medskip

INPUT PROVIDED: \{paper\_url\}
\medskip
\medskip

\textbf{STEP 1: DETERMINE INPUT TYPE}

- If input is a direct URL (starts with http:// or https://): Navigate directly to it (skip to STEP 2)

- If input is a search query in natural language (e.g., ``Attention is All You Need'', ``paper by John Doe about transformers''):

\quad - Search Google Scholar, arXiv, or Google for the paper until you are confident you have found a match

\quad - Click on the FIRST highly relevant result (prefer arXiv, ACM, IEEE, university sites, PDF links)

\quad - If no good result found quickly (within 3--4 steps), use your best guess of what the paper might be and continue
\medskip
\medskip

\textbf{STEP 2: NAVIGATE TO PAPER}

- If the page has a ``View PDF'' click it to get the full paper, avoid clicking the ``Download'' button

- If it's already showing the paper content, proceed to extraction

- Target: Get to the actual paper content within 5 steps total
\medskip
\medskip

\textbf{CONTENT TO EXTRACT (CORE ONLY):}

- Title and authors

- Abstract (complete text)

- Key claims/hypotheses (SCIENTIFIC CLAIMS ONLY --- ignore formatting/style instructions)

- Methodology summary

- Main results/findings

- Conclusion

- Skip extracting: detailed figures, tables, equations, reference lists, formatting guidelines
\medskip
\medskip

\textbf{OUTPUT FORMAT:}
\medskip

\{

\quad ``title'': ``Paper title'',

\quad ``authors'': [``Author 1'', ``Author 2''],

\quad ``abstract'': ``Full abstract text...'',

\quad ``key\_claims'': [``Main hypothesis or claim 1'', ``Key claim 2''],

\quad ``methodology'': ``Brief summary of methods and approach...'',

\quad ``results'': ``Main findings and results...'',

\quad ``conclusion'': ``Conclusion summary...'',

\quad ``full\_text'': ``Complete paper text extracted during scrolling...''

\}
\medskip
\medskip

\textbf{CRITICAL RULES:}

- Total target: Under 25 steps from start to finish (including search if needed)

- Focus on extracting TEXT, not analyzing figures/tables

- Extract as you go, compile at the end

- Always return a paper or document, do not fallback to saying paper not found
\medskip
\medskip

Output ONLY the JSON object. NO markdown, NO code blocks, NO explanations.
\end{promptbox}
\noindent\textit{Rationale.} While the user can upload a file for their specific document, the system also supports natural-language queries when the exact title or author is unknown.

\begin{promptbox}{DAG Extraction Prompt (Part 1/3): Instructions}
Your name is Plato, you are a precise information extraction system that analyzes academic text and structures it as a directed acyclic graph (DAG) of scientific claims and evidence built by researchers at the University of Florida.
\smallskip

Your task is to extract the major statements and claims from the provided text and connect them based on logical relationships with rich semantic roles.
\smallskip

\textbf{EXHAUSTIVENESS REQUIREMENTS:}
\begin{itemize}
  \item Extract the major SCIENTIFIC statements from the text (hypotheses, evidence, methods, results)
  \item Break down complex statements into smaller, self-contained claims
  \item Each node should represent one clear statement
  \item LIMIT: Maximum of \{max\_nodes\} nodes total (including the hypothesis)
  \item Prefer larger, comprehensive nodes over many tiny ones
\end{itemize}

\textbf{EXCLUDE THE FOLLOWING (DO NOT extract these as nodes):}
\begin{itemize}
  \item Formatting instructions (APA style, citation format, page layout, margins, fonts)
  \item Style guidelines (headings, indentation, spacing)
  \item Structural requirements (abstract length, section order)
  \item Writing tips or general advice
  \item Meta-commentary about the paper itself
\end{itemize}

FOCUS ONLY ON: Scientific claims, hypotheses, evidence, methodology, results, conclusions
\smallskip

\textbf{GRAPH CONSTRUCTION RULES:}
\begin{itemize}
  \item Create a strictly directed acyclic graph (DAG) structure
  \item The HYPOTHESIS (main research question or claim) MUST ALWAYS be the root node (ID 0)
  \item Children flow from parents: Evidence supports Claims, Claims support Hypothesis, etc.
  \item Use parent/child relationships to show logical dependencies
  \item Each node must specify its role in the argument structure
\end{itemize}

\textbf{NODE ROLES (choose ONE per node):}
\begin{itemize}
  \item Hypothesis: The main research hypothesis or central claim (MUST be ID 0, the root)
  \item Conclusion: Final conclusions drawn from the research
  \item Claim: An assertion or statement made in the paper
  \item Evidence: Data, observations, or results that support claims
  \item Method: Description of methodology, techniques, or procedures
  \item Result: Specific findings or outcomes from experiments/analysis
  \item Assumption: Underlying assumptions or premises
  \item Counterevidence: Evidence that contradicts or challenges claims
  \item Limitation: Acknowledged limitations or constraints
  \item Context: Background information or related work
\end{itemize}

\textbf{OUTPUT FORMAT REQUIREMENTS:}
\begin{itemize}
  \item Output ONLY valid JSON with no additional commentary or explanation
  \item Use exactly one key: ``nodes'' (no separate edges --- relationships are in parent/child fields)
  \item No extra keys, no markdown formatting, no code blocks
  \item Do NOT truncate the output --- if needed, make node text shorter to fit more nodes
\end{itemize}

\textbf{CRITICAL JSON RULES (to ensure valid parsing):}
\begin{itemize}
  \item Convert all LaTeX notation to plain text (e.g., ``\$\textbackslash mathcal\{D\}\$'' becomes ``dataset D'')
  \item Replace all mathematical symbols with words (e.g., ``$\alpha$'' becomes ``alpha'', ``$\sum$'' becomes ``sum'')
  \item Never include backslashes in text fields except valid JSON escapes: \textbackslash n, \textbackslash t, \textbackslash ", \textbackslash\textbackslash
  \item Remove or describe all special formatting from the original paper
  \item All text must be valid JSON string content --- no unescaped special characters
\end{itemize}
\end{promptbox}
\noindent\textit{Prompt metadata.} Part 1/3 defines instruction and constraints. The max number of nodes is set by \{max\_nodes\}.

\begin{promptbox}{DAG Extraction Prompt (Part 2/3): Required Schema}
\textbf{JSON Structure:}
\medskip

\{

\quad ``nodes'': [

\quad\quad \{

\quad\quad\quad ``id'': 0,

\quad\quad\quad ``text'': ``Main hypothesis or research question'',

\quad\quad\quad ``role'': ``Hypothesis'',

\quad\quad\quad ``parents'': null,

\quad\quad\quad ``children'': [1, 2, 3]

\quad\quad \},

\quad\quad \{

\quad\quad\quad ``id'': 1,

\quad\quad\quad ``text'': ``A claim that supports the hypothesis'',

\quad\quad\quad ``role'': ``Claim'',

\quad\quad\quad ``parents'': [0],

\quad\quad\quad ``children'': [4, 5]

\quad\quad \},

\quad\quad \{

\quad\quad\quad ``id'': 2,

\quad\quad\quad ``text'': ``Evidence supporting a claim'',

\quad\quad\quad ``role'': ``Evidence'',

\quad\quad\quad ``parents'': [1],

\quad\quad\quad ``children'': null

\quad\quad \}

\quad ]

\}
\medskip

\textbf{FIELD REQUIREMENTS:}
\begin{itemize}
  \item id: Sequential integer starting from 0
  \item text: Clear, concise description of the statement (string)
  \item role: ONE of the roles listed above (string)
  \item parents: List of parent node IDs [int, ...] or null if root (must be null ONLY for ID 0)
  \item children: List of child node IDs [int, ...] or null if leaf node
\end{itemize}

\textbf{VALIDATION CHECKLIST:}
\begin{itemize}
  \item ID 0 is ALWAYS the Hypothesis (the root node)
  \item ID 0 has parents: null
  \item All non-root nodes have parents: [list of IDs]
  \item Leaf nodes have children: null
  \item Non-leaf nodes have children: [list of IDs]
  \item All node IDs are sequential starting from 0
  \item All referenced parent/child IDs exist in the graph
  \item No cycles exist (child IDs are always greater than parent IDs)
  \item Each node has exactly ``id'', ``text'', ``role'', ``parents'', ``children'' fields
  \item Role is one of: Hypothesis, Conclusion, Claim, Evidence, Method, Result, Assumption, Counterevidence, Limitation, Context
\end{itemize}
\end{promptbox}
\noindent\textit{Prompt metadata.} Part 2/3 specifies the required JSON schema and validation constraints.

\begin{promptbox}{DAG Extraction Prompt (Part 3/3): Examples}

\textbf{Example for a simple paper about machine learning:}
\medskip

\{

\quad ``nodes'': [

\quad\quad \{

\quad\quad\quad ``id'': 0,

\quad\quad\quad ``text'': ``Neural networks can improve image classification accuracy'',

\quad\quad\quad ``role'': ``Hypothesis'',

\quad\quad\quad ``parents'': null,

\quad\quad\quad ``children'': [1, 2]

\quad\quad \},

\quad\quad \{

\quad\quad\quad ``id'': 1,

\quad\quad\quad ``text'': ``Convolutional layers extract hierarchical features from images'',

\quad\quad\quad ``role'': ``Claim'',

\quad\quad\quad ``parents'': [0],

\quad\quad\quad ``children'': [3]

\quad\quad \},

\quad\quad \{

\quad\quad\quad ``id'': 2,

\quad\quad\quad ``text'': ``Our CNN achieved 95\% accuracy on ImageNet'',

\quad\quad\quad ``role'': ``Result'',

\quad\quad\quad ``parents'': [0],

\quad\quad\quad ``children'': null

\quad\quad \},

\quad\quad \{

\quad\quad\quad ``id'': 3,

\quad\quad\quad ``text'': ``We used backpropagation to train the network'',

\quad\quad\quad ``role'': ``Method'',

\quad\quad\quad ``parents'': [1],

\quad\quad\quad ``children'': null

\quad\quad \}

\quad ]

\}
\medskip

\textbf{TEXT TO ANALYZE:} \{raw\_text\}
\smallskip

Remember: Output ONLY the JSON object. No explanations, no markdown, no code blocks.
\end{promptbox}
\noindent\textit{Prompt metadata.} Part 3/3 provides JSON examples followed by the raw text placeholder.

\FloatBarrier

\section{Hyperparameter Settings}

For reproducibility, we summarize the default scorer configuration available in the code release. Unless otherwise stated, all scalar values are in $[0,1]$ and all scores are clipped into $[0,1]$ after computation. Appendix~\ref{app:scoring_math} provides the exact scoring equations.
Each node receives six verification metrics (each in $[0,1]$): credibility, relevance, evidence strength, method rigor, reproducibility, and citation support. A global per-metric weight (all default to 1.0) is applied before role-specific mixing. Node quality is computed using the role-specific weights in Table~\ref{tab:node_quality_weights} (normalized by $\ell_1$ magnitude). Raw edge confidence is a weighted mixture of five features; the paper-level score aggregates six graph-level components. Table~\ref{tab:defaults} summarizes all default weights.

\begin{table}[!htbp]
\centering
\caption{Default weights for edge features and paper-level components.}
\label{tab:defaults}
\small
\begin{tabularx}{\columnwidth}{X c @{\hspace{2em}} X c}
\toprule
\textbf{Edge feature} & \textbf{Wt.} & \textbf{Paper component} & \textbf{Wt.} \\
\midrule
role prior    & 0.30 & Bridge      & 0.25 \\
parent qual   & 0.20 & Best Path   & 0.25 \\
child qual    & 0.20 & Redundancy  & 0.15 \\
alignment     & 0.10 & Fragility   & $-$0.15 \\
synergy       & 0.20 & Coherence   & 0.10 \\
              &      & Coverage    & 0.10 \\
\bottomrule
\end{tabularx}
\end{table}

\begin{table*}[!htbp]
\centering
\caption{Default role-specific node-quality weights. Columns correspond to (credibility, relevance, evidence strength, method rigor, reproducibility, citation support).}
\label{tab:node_quality_weights}
\small
\begin{tabular*}{\textwidth}{@{\extracolsep{\fill}}lcccccc}
\toprule
Role & cred. & rel. & evid. & rigor & repr. & cite \\
\midrule
Hypothesis      & 0.50 & 0.50 & 0.00 & 0.00 & 0.00 & 0.00 \\
Conclusion      & 0.60 & 0.40 & 0.00 & 0.00 & 0.00 & 0.00 \\
Claim           & 0.45 & 0.35 & 0.20 & 0.00 & 0.00 & 0.00 \\
Evidence        & 0.20 & 0.00 & 0.50 & 0.00 & 0.00 & 0.30 \\
Method          & 0.10 & 0.00 & 0.00 & 0.60 & 0.30 & 0.00 \\
Result          & 0.40 & 0.30 & 0.30 & 0.00 & 0.00 & 0.00 \\
Assumption      & 0.60 & 0.40 & 0.00 & 0.00 & 0.00 & 0.00 \\
Counterevidence & 0.20 & 0.00 & 0.50 & 0.00 & 0.00 & 0.30 \\
Limitation      & 0.50 & 0.50 & 0.00 & 0.00 & 0.00 & 0.00 \\
Context         & 0.40 & 0.60 & 0.00 & 0.00 & 0.00 & 0.00 \\
\bottomrule
\end{tabular*}
\end{table*}

Trust propagation and edge gating are controlled by a small set of interpretable hyperparameters:

\begin{table*}[!htbp]
\begin{center}
\begin{tabular*}{\textwidth}{@{\extracolsep{\fill}}lll}
\toprule
Parameter & Default & Meaning \\
\midrule
\texttt{enabled} & True & enable/disable trust propagation \\
\texttt{agg} & \texttt{min} & parent aggregation (\texttt{min}, \texttt{mean}, \texttt{softmin}, \texttt{dampmin}) \\
$\alpha$ & 1.0 & exponent on parent trust (negative values are clamped to 0) \\
$\eta$ & $2^{-1/8}\approx 0.917$ & floor on the edge-gating factor \\
$\beta$ & 6.0 & \texttt{softmin} sharpness (higher $\Rightarrow$ closer to \texttt{min}) \\
$\lambda$ & 0.35 & \texttt{dampmin} mixing (higher $\Rightarrow$ closer to \texttt{mean}) \\
\texttt{default\_raw\_conf} & 0.5 & fallback raw confidence if missing \\
\bottomrule
\end{tabular*}
\end{center}
\end{table*}

\subsection{Notes}
The default $\eta$ is intentionally high: it prevents complete collapse of confidence when upstream trust is incomplete during real-time streaming. If you want propagation to be more punitive, decrease $\eta$ and/or increase $\alpha$.

\section{Scoring Mathematics and Design Rationale}
\label{app:scoring_math}

This section outlines the equations for the scorer implementation (\texttt{kg\_realtime\_scoring.py}) and covers the design choices that do not appear in the main paper for readability.

Let $v\in V$ denote a node with role $\rho_v$, text $t_v$, and six verification metrics $\mathbf{m}_v\in[0,1]^6$ with components $m_{v,k}$. Let $q_v\in[0,1]$ be the node quality, $t_v\in[0,1]$ the propagated trust, and $C^{\text{raw}}_{u\to v},\,C_{u\to v}\in[0,1]$ the raw final edge confidences.

Global metric weights $g_k\ge 0$ are applied to each metric, with clipping back into $[0,1]$:
\begin{equation}
\tilde{m}_{v,k} = \mathrm{clip}_{[0,1]}\!\left(g_k\,m_{v,k}\right).
\end{equation}
Role-specific node-quality weights $w^{(\rho_v)}_k$ are then applied and normalized by $\ell_1$ magnitude:
\begin{equation}
q_v = \mathrm{clip}_{[0,1]}\!\left(\frac{\sum_k w^{(\rho_v)}_k\,\tilde{m}_{v,k}}{\sum_k |w^{(\rho_v)}_k| + \epsilon}\right),
\end{equation}
where $\epsilon$ is a small constant for numerical safety. If a role has no specified weights, the implementation falls back to the simple mean of the six metrics.

$\ell_1$ normalization makes the quality score invariant to overall weight scale and simplifies ablations. Changing a role's weights changes \emph{relative emphasis} rather than raw magnitude.

Edges $(u\to v)$ are described by five interpretable features:
\begin{itemize}
    \item role prior $r_{\rho_u\to\rho_v}\in[0,1]$ (defaults to 0.5 if unspecified),
    \item parent and child qualities $q_u,q_v$,
    \item lexical alignment $a_{u,v}=\mathrm{Jaccard}(\mathrm{tok}(t_u),\mathrm{tok}(t_v))$,
    \item role-pair synergy $s_{u,v}\in[0,1]$ (below).
\end{itemize}

Raw confidence is a clipped weighted sum:
\begin{equation}
\begin{aligned}
C^{\text{raw}}_{u\to v}
&=\mathrm{clip}_{[0,1]}\!\Big(
\theta_r r_{\rho_u\to\rho_v} + \theta_p q_u + \theta_c q_v \\
&\hspace{2.2em} + \theta_a a_{u,v} + \theta_s s_{u,v}
\Big).
\end{aligned}
\end{equation}

For select role pairs $(\rho_u,\rho_v)$, synergy mixes the parent and child metric vectors with role-pair-specific weights $\mathbf{w}^{\text{par}}_{\rho_u\rho_v}$ and $\mathbf{w}^{\text{chi}}_{\rho_u\rho_v}$, then averages the two mixes:
\begin{equation}
\begin{aligned}
s_{u,v}
&= \mathrm{clip}_{[0,1]}\!\Big(
\tfrac{1}{2}\,\mathrm{mix}(\tilde{\mathbf{m}}_u,\mathbf{w}^{\text{par}}_{\rho_u\rho_v})\\
&\hspace{2.2em}+\tfrac{1}{2}\,\mathrm{mix}(\tilde{\mathbf{m}}_v,\mathbf{w}^{\text{chi}}_{\rho_u\rho_v})
\Big).
\end{aligned}
\end{equation}
\begin{equation}
\mathrm{mix}(\mathbf{x},\mathbf{w})
=\frac{\sum_k w_k x_k}{\sum_k |w_k| + \epsilon}.
\end{equation}
If a role pair is unspecified, the default is an equal-weight average of the parent and child metric means.

This update can be interpreted as passing scalar ``trust messages'' along edges, analogous in spirit to BP/sum-product message passing but replacing probabilistic marginals with verification-derived trust and applying a gate to control downstream influence. \cite{pearl1988probabilistic, kschischang2001factor, yedidia2003understanding, wainwright2008variational}

If $v$ has no parents, trust equals quality: $t_v=q_v$. Otherwise, each parent $u\in\mathrm{Pa}(v)$ contributes
\begin{equation}
z_{u\to v} = \big(\max(10^{-6},t_u)\big)^\alpha \, C^{\text{raw}}_{u\to v}.
\end{equation}
These contributions are aggregated to a single support value $\mathrm{Agg}(\{z_{u\to v}\})$:
\begin{equation}
t_v=\mathrm{clip}_{[0,1]}\!\left(q_v \cdot \mathrm{Agg}(\{z_{u\to v}\}_{u\in\mathrm{Pa}(v)})\right).
\end{equation}

Let {$Z=\{z_i\}_{i=1}^n$} be the parent contributions for a fixed child node. Four aggregation modes are supported:
\begin{equation}
\mathrm{Agg}(Z)=\begin{cases}
\min_i z_i & \texttt{min}\\[4pt]
\frac{1}{n}\sum_i z_i & \texttt{mean}\\[4pt]
\frac{\sum_i w_i z_i}{\sum_i w_i},\; w_i\!=\!\exp\!\bigl(\!-\beta\,\tfrac{z_i-z_{\min}}{z_{\max}-z_{\min}}\bigr) & \texttt{softmin}\\[4pt]
(1\!-\!\lambda)\min_i z_i + \lambda\,\mathrm{Agg}_{\texttt{mean}}(Z) & \texttt{dampmin}
\end{cases}
\end{equation}
where $\beta=\texttt{softmin\_beta}$ (default 6.0) and $\lambda=\texttt{dampmin\_lambda}$ (default 0.35). If $z_{\max}-z_{\min}\le 10^{-12}$ or $\beta=0$, \texttt{softmin} falls back to \texttt{mean}.

Regarding our design choice, separating $C^{\text{raw}}_{u\to v}$ from trust makes the system easier to debug: raw confidence isolates local plausibility, while trust captures global propagation effects.

\subsection{Final (gated) edge confidence}

After trust is computed, raw confidence is gated by parent trust with floor $\eta$:
\begin{equation}
C_{u\to v}=\mathrm{clip}_{[0,1]}\!\left(C^{\text{raw}}_{u\to v}\,\big(\eta + (1-\eta)\,t_u^\alpha\big)\right).
\end{equation}

As for design choices, the floor $\eta$ prevents propagation from over-penalizing graphs early in the streaming process; lowering $\eta$ makes the system more ``all-or-nothing.''

\subsection{Graph-level components}

Let $H$ be hypothesis nodes and $C$ be conclusion nodes. Define the \emph{bridge} node set as those reachable from some $h\in H$ and that can reach some $c\in C$. All graph-level components are computed on this bridge:

\begin{itemize}
    \item Bridge coverage: $\frac{|\mathrm{BridgeNodes}|}{|V|}$.
    \item Best-path reliability: among all hypothesis$\rightarrow$conclusion paths, choose the path that maximizes the product of edge confidences, and report the geometric mean along that path (length-normalized).
    \item Redundancy: unit-capacity max-flow between a super-source connected to hypotheses and a super-sink connected from conclusions; mapped as $\min(\mathrm{flow}/3,\,1)$.
    \item Fragility: min-cut on the same construction with capacities $\max(10^{-6},\,1-C_{u\to v})$, normalized by the number of bridge edges.
    \item Coherence: fraction of bridge edges with role prior $\ge 0.5$.
    \item Coverage: fraction of key roles $\{\textbf{Evidence},\textbf{Method},\textbf{Result}\}$ present on the bridge.
\end{itemize}

A few design choices warrant discussion:
(i) ``Bridge-only'' computation reduces cost and avoids rewarding disconnected subgraphs.
(ii) \textbf{Best-path reliability is treated as the primary \emph{argument strength} signal.}
Among all hypothesis$\rightarrow$conclusion paths, we select the path that maximizes the \emph{product} of gated edge confidences and then report a \emph{geometric mean} (length-normalized) along that path.
This yields an auditable chain for inspection and reduces systematic bias toward shorter paths.
\begin{equation}
\begin{aligned}
P^\star &\in \arg\max_{P\in\mathcal{P}(H,C)} \prod_{(u\to v)\in P} C_{u\to v},\\
R_{\text{best}} &= \left(\prod_{(u\to v)\in P^\star} C_{u\to v}\right)^{1/|P^\star|}.
\end{aligned}
\end{equation}
(iii) Redundancy uses unit-capacity max-flow and is softly capped at 3 to reflect diminishing returns.

\subsection{Final score normalization}

Let the six component values be $s_1,\dots,s_6\in[0,1]$ and weights be $\gamma_1,\dots,\gamma_6$ (weights may be negative to express penalties). The raw score is:
\begin{equation}
s_{\mathrm{raw}}=\sum_{i=1}^{6}\gamma_i s_i.
\end{equation}
To map into $[0,1]$ even with negative weights, we compute $p=\sum_{i:\gamma_i>0}\gamma_i$ and $n=\sum_{i:\gamma_i<0}\gamma_i$ and set
\begin{equation}
s_{[0,1]}=\mathrm{clip}_{[0,1]}\!\left(\frac{s_{\mathrm{raw}}-n}{p-n+\epsilon}\right).
\end{equation}
Finally, to avoid exact endpoints (useful for calibration and downstream learning), we map to $(0,1)$ with a small $\varepsilon$:
\begin{equation}
s=\varepsilon + (1-2\varepsilon)\,s_{[0,1]}.
\end{equation}

\subsection{Role-transition priors (non-default entries)}

Unspecified role transitions default to 0.5. Table~\ref{tab:role_priors} lists the non-default priors used in the code release.

\begin{table}[!htbp]
\centering
\caption{Non-default role transition priors $r_{\rho_u\to\rho_v}$}
\label{tab:role_priors}
\small
\begin{tabularx}{\linewidth}{X X c}\toprule
Parent role & Child role & Prior \\
\midrule
Hypothesis & Claim & 0.75 \\
Hypothesis & Evidence & 0.75 \\
Hypothesis & Method & 0.50 \\
Hypothesis & Result & 0.25 \\
Hypothesis & Conclusion & 0.25 \\
Evidence & Result & 1.00 \\
Evidence & Claim & 0.50 \\
Evidence & Conclusion & 0.75 \\
Method & Result & 0.75 \\
Method & Evidence & 0.50 \\
Result & Conclusion & 0.75 \\
Claim & Conclusion & 0.50 \\
Claim & Evidence & 0.50 \\
Context & Claim & 0.50 \\
Assumption & Claim & 0.50 \\
Assumption & Method & 0.50 \\
Counterevidence & Claim & 0.75 \\
Counterevidence & Conclusion & 0.75 \\
\bottomrule
\end{tabularx}
\end{table}

\subsection{Role-pair synergy specifications (defaults)}

Table~\ref{tab:synergy_specs} lists the role pairs with explicit synergy mixes. Unspecified pairs fall back to equal-weight parent/child mean-metrics mixing.

\begin{table}[!htbp]
\centering
\caption{Role-pair synergy specifications (parent/child metric mixes)}
\label{tab:synergy_specs}
\small
\begin{tabularx}{\columnwidth}{X X c}
\toprule
Role pair $(\rho_u\!\to\!\rho_v)$ & Parent mix & Child mix \\
\midrule
Evi $\to$ Clm & evid:.5, cite:.3, cred:.2 & cred:.6, rel:.4 \\
Evi $\to$ Res & evid:.5, cite:.3, cred:.2 & cred:.5, rel:.5 \\
Evi $\to$ Con & evid:.5, cite:.4, cred:.1 & cred:.7, rel:.3 \\
Met $\to$ Res & rigor:.6, repr:.3, cred:.1 & cred:.6, rel:.4 \\
Hyp $\to$ Clm & cred:.3, rel:.7 & cred:.6, rel:.4 \\
Hyp $\to$ Evi & cred:.4, rel:.6 & cred:.5, rel:.5 \\
Clm $\to$ Con & cred:.6, rel:.4 & cred:.7, rel:.3 \\
\bottomrule
\end{tabularx}
\end{table}

\subsection{Interpretation and Graph Construction Algorithm}
Synergy allows edge confidence to reflect whether the \emph{combination} of a parent and child role is well-supported according to the metrics that matter most for that relationship (e.g., Evidence$\to$Claim emphasizes evidence strength and citation support on the parent, but credibility on the child). Algorithm~\ref{alg:dag_construction} shows the pseudocode for DAG extraction with validation.

\begin{algorithm}[H]
\caption{DAG Construction with Validation}
\label{alg:dag_construction}
\begin{algorithmic}[1]
\STATE \textbf{Input:} Paper text $P$, ontology $\mathcal{R}$
\STATE \textbf{Output:} Validated DAG $\mathcal{G} = (V, E)$
\STATE
\STATE $\mathcal{G} \gets $ LLM\_Extract($P, \mathcal{R}$)
\STATE $\text{valid} \gets \text{False}$
\STATE $\text{attempts} \gets 0$
\WHILE{not valid and attempts $< 3$}
    \IF{HasCycle($\mathcal{G}$)}
        \STATE feedback $\gets$ ``Graph contains cycles''
    \ELSIF{ViolatesRoleConstraints($\mathcal{G}$)}
        \STATE feedback $\gets$ ``Invalid role transitions detected''
    \ELSE
        \STATE valid $\gets$ True
    \ENDIF
    \IF{not valid}
        \STATE $\mathcal{G} \gets $ LLM\_Refine($P, \mathcal{G}, \text{feedback}$)
        \STATE attempts $\gets$ attempts $+ 1$
    \ENDIF
\ENDWHILE
\RETURN $\mathcal{G}$
\end{algorithmic}
\end{algorithm}

\vspace{10em}
\section{Role Ontology Specification}
\begin{table}[!htbp]
\centering
\caption{Semantic role ontology. Each node must be assigned exactly one role.}
\label{tab:role_ontology}
\small
\begin{tabularx}{\columnwidth}{l X}
\toprule
\textbf{Role} & \textbf{Definition} \\
\midrule
Hypothesis & Testable proposition or research question the paper investigates. \\
Evidence & Factual observations, data, or citations from prior work that support claims. \\
Claim & Assertions derived from evidence or reasoning. \\
Method & Descriptions of procedures, algorithms, or experimental designs. \\
Result & Outcomes obtained from applying Methods. \\
Conclusion & High-level takeaways synthesizing Results and answering Hypotheses. \\
Assumption & Unstated premises or idealizations underlying the research. \\
Counterevidence & Data or observations contradicting Claims or Results. \\
Limitation & Acknowledged weaknesses, scope restrictions, or threats to validity. \\
Context & Background information, motivation, or domain knowledge framing the research. \\
\bottomrule
\end{tabularx}
\end{table}





\end{document}